\documentclass[10pt]{cai}

\begin{document}
\def\conferenceyear{2025}
\begin{center}

\title{Evaluating the Effectiveness of Cost-Efficient \\ Large Language Models in Benchmark Biomedical Tasks}
\maketitle

\thispagestyle{empty}

\begin{tabular}{cc}
Israt Jahan,
Md Tahmid Rahman Laskar, Chun Peng,
Jimmy Xiangji Huang\upstairs{*} \\
{\small York University} \\ {\small Toronto, Ontario, Canada} 

\end{tabular}

\emails{
  \upstairs{*}jhuang@yorku.ca 
}
\vspace*{0.2in}
\end{center}

\begin{abstract} 
This paper presents a comprehensive evaluation of cost-efficient Large Language Models (LLMs) for diverse biomedical tasks spanning both text and image modalities. We evaluated a range of closed-source and open-source LLMs on tasks such as biomedical text classification and generation, question answering, and multimodal image processing. Our experimental findings indicate that there is no single LLM that can consistently outperform others across all tasks. Instead, different LLMs excel in different tasks. While some closed-source LLMs demonstrate strong performance on specific tasks, their open-source counterparts achieve comparable results (sometimes even better), with additional benefits like faster inference and enhanced privacy. Our experimental results offer valuable insights for selecting models that are optimally suited for specific biomedical applications.
\end{abstract}

\section{Introduction}
Large Language Models (LLMs) have demonstrated impressive capabilities across various domains \cite{laskar-etal-2023-systematic}, including biomedicine \cite{jahan2024comprehensive}. 
Recently, the capability of LLMs from only understanding textual data has been further extended, allowing them to understand multimodal data \cite{tian2024opportunities}. These improved capabilities of LLMs have made it possible to utilize them in various real-world biomedical applications. In biomedicine, LLMs have the potential to perform critical tasks like drug discovery, disease diagnosis, radiology report generation, etc \cite{jahan2024comprehensive,tian2024opportunities}. 

However, despite the potential of AI to revolutionize biomedicine, the utilization of LLMs in real-world biomedical settings is very limited
due to the high cost (e.g., computing resources, API cost, data annotation) associated with LLM development and deployment \cite{tian2024opportunities}. Moreover, sharing sensitive data externally for model development and inference raises privacy concerns, necessitating secure pipelines, which further increases costs. 

To this end, this paper aims to study how to make LLMs more efficient and cost-effective while retaining their accuracy in performing diverse biomedical tasks in practical scenarios. This would require an extensive evaluation of the smaller LLMs that are currently available to study their capabilities and limitations in benchmark biomedical datasets and tasks. Our hypothesis is that while larger LLMs may generally exhibit superior performance, strategically chosen smaller LLMs can offer a compelling balance of performance and efficiency.

By benchmarking the performance of the cost-efficient open-source and closed-source models, this paper makes the following key contributions: 

\begin{itemize}
    \item 
 For open-source models, this would give insights on which models to select for further fine-tuning to make them more specialized in certain biomedical tasks. 
        \item  For closed-source models, in addition to identifying which one of them can be used in practical applications via their respective APIs, our findings will also be useful to select the right closed-source model for the development of specialized open-source models for biomedicine (i.e., using closed-source models for generating synthetic data for continual pre-training or instruction tuning of the open-source models).
        \item The findings from this research will provide valuable insights for researchers and practitioners seeking to deploy these models in the biomedical domain.
\end{itemize}

\begin{figure*}
    \centering
    \includegraphics[width=0.85\linewidth]{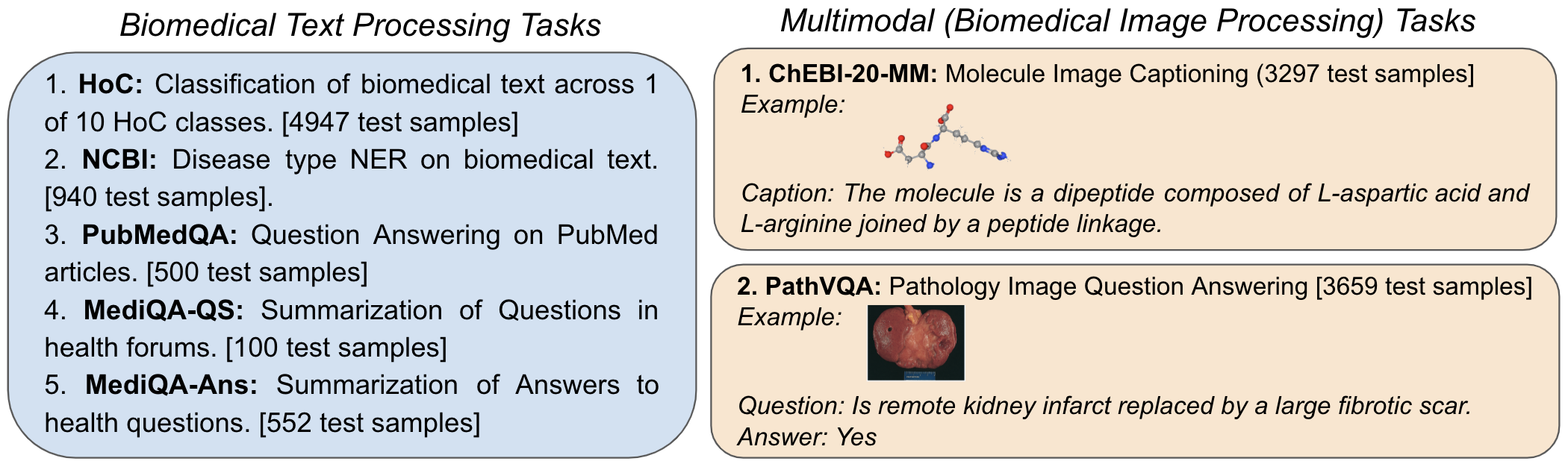}
    \caption{An overview of the biomedical tasks evaluated in this paper. The images for ChEBI-20-MM and PathVQA are taken from Liu et al. \cite{liu2024scientificmoleculechebi} and He et al. \cite{he2020pathvqa}.}
    \label{fig:overview}
\end{figure*}

\section{Related Work}
In recent years, the utilization of pre-trained transformer-based models by fine-tuning on task-specific biomedical datasets have demonstrated state-of-the-art performance in a wide range of biomedical text processing tasks \cite{jahan2024comprehensive}. However, one major limitation of using such pre-trained biomedical models is that they do not have instruction-following capability and also require task-specific large annotated datasets for fine-tuning, which is significantly less available in the biomedical domain \cite{jahan2024comprehensive}. In this regard, having a strong zero-shot model with instruction-following capability could potentially alleviate the need for large annotated datasets, enabling the model to perform well in tasks that it was not explicitly trained on.

While prior research has demonstrated that LLMs can outperform the state-of-the-art fine-tuned biomedical models even in zero-shot scenarios on biomedical datasets that have smaller training sets \cite{jahan2024comprehensive}, the evaluation was predominantly focused on earlier generation LLMs (e.g., GPT-3.5 \cite{openai2023gpt4}, Claude-2 \cite{anthropicclaude3}, LLaMA-2-13B \cite{touvron2023llama2}, and PaLM-2 \cite{anil2023palm2}). Moreover, there is a lack of comprehensive evaluation of LLMs in multimodal biomedical tasks. While numerous newly proposed LLMs have demonstrated multimodal capabilities, a comprehensive benchmarking of these new LLMs in biomedicine across multimodal tasks is still missing \cite{tian2024opportunities}.  In addition, prior research has also ignored the importance of computational efficiency which is a crucial factor for practical deployment of LLMs \cite{fu2024tiny}. 

To address the above issues, in this paper, we provide a comprehensive evaluation of efficient LLMs that require lower computational resources. Our extensive experiments of these cost-effective LLMs across diverse biomedical tasks (both text data and images) would provide critical insights on their applicability in real-world clinical settings. 
\section{Methodology}
\subsection{Datasets and Tasks}

For evaluation, we use biomedical tasks from diverse modalities (also see Figure \ref{fig:overview}): 

\noindent{\textbf{(i) Tasks for Biomedical Text Data:}} This category consists of the tasks that require the analysis of biomedical text data. We use the Hallmarks of Cancer \cite{hoc} dataset for \textit{biomedical text classification} across 1 of the 10 HoC classes, the NCBI-disease \cite{ncbi} named entity recognition (NER) dataset for \textit{biomedical entity extraction} (disease entity), the PubMedQA \cite{jin2019pubmedqa} dataset for \textit{biomedical question answering}, the MediQA-QS \cite{abacha2021overview} dataset for \textit{medical question summarization} and the MediQA-ANS \cite{savery2020questionmediqaans} dataset for \textit{medical answer summarization.}

\noindent{\textbf{(ii) Tasks for Biomedical Image Data (Multimodal):}} We use the ChEBI-20-MM \cite{liu2024scientificmoleculechebi} dataset for \textit{caption generation from molecular images} and the PathVQA \cite{he2020pathvqa} dataset for \textit{question answering from pathology images}. For PathVQA, we use its binary \textit{Yes/No} type question answering subset since the other subset that requires open-ended answer generation is quite similar to the caption generation task.

\subsection{Prompt Construction}
\label{sec:prompt}
Prompts are essential for interacting with LLMs. For biomedical text processing, we use prompts from Jahan et al. \cite{jahan2024comprehensive}. In biomedical image processing, tasks often require 
minimal prompt engineering. For example, a simple prompt—\texttt{``Generate a descriptive caption of the molecular structure image''}—works well for molecular captioning, which we also use. In PathVQA, we use dataset-provided questions as the prompt. 

\subsection{Models}
We primarily use the cost-efficient LLMs currently available, considering their real-world applicability. Therefore, for closed-source LLMs, we use: 
(i) GPT-4o-Mini \cite{openai2023gpt4}, (ii) Gemini-1.5-Flash \cite{team2023gemini}, (iii) Claude-3-Haiku \cite{anthropicclaude3}. All these closed-source models have multimodal capabilities.
For the open-source LLMs, we select models having fewer than 13B parameters. For text-based tasks, we select the instruction-tuned version of respective open-source models such that they can properly follow the instructions in the prompt: (iv) LLaMA-3.1-8B-Instruct \cite{dubey2024llama3}, (v) Qwen-2.5-7B-Instruct \cite{yang2024qwen2.5}, (vi) Mistral-7B-v0.3-Instruct \cite{jiang2023mistral}, and (vii) Phi-3.5-Mini-3.8B-Instruct \cite{abdin2024phi}.
With the recent success of reasoning-based LLMs like DeepSeek-R1 \cite{guo2025deepseek}, we also use its distilled versions based on Qwen and LLaMA, (vii) DeepSeek-R1-Distill-Qwen-7B and (viii) DeepSeek-R1-Distill-LLaMA-8B, respectively. For image-based tasks using open-source models, we select: Phi-3.5-Vision \cite{abdin2024phi}, Qwen-2-VL \cite{wang2024qwen2vl}, LLaVA-Next \footnote{\url{https://llava-vl.github.io/blog/2024-01-30-llava-next/}} based on Mistral-7B \cite{jiang2023mistral}, Janus-Pro \cite{chen2025janus}, and LLaMA-3.2-11B-Vision\footnote{\url{https://huggingface.co/meta-llama/Llama-3.2-11B-Vision}}.

The inference of each model was conducted by leveraging zero-shot prompts (as described in Section \ref{sec:prompt}) on a machine with 1 NVIDIA A100 GPU. The temperature value was set to 1.0, with other decoding parameters being set to the default values in the respective API providers for the closed-source models and in HuggingFace\footnote{\url{https://huggingface.co/}} for the open-source models. 
\subsection{Evaluation}
For \textbf{classification} and \textbf{information extraction} tasks, a parsing script is required to first extract answers from the LLM-generated responses to compare against gold labels \cite{laskar-etal-2024-systematic}. Afterwards, their performance is measured using dataset-specific metrics like \textit{Accuracy}, \textit{Precision}, \textit{Recall}, and \textit{F1}, which are commonly used in the literature \cite{jahan2024comprehensive}.

For \textbf{generative} tasks (e.g., summarization or caption generation), parsing scripts are not required \cite{laskar-etal-2024-systematic} and the full response generated by LLMs are compared against the gold reference. Similar to prior research \cite{jahan2024comprehensive}, we use \textit{ROUGE} \cite{rouge} and \textit{BERTScore} \cite{zhang2019bertscore} metrics. 
\section{Results and Discussion}

\begin{table}[t]
\tiny
\centering
\begin{tabular}{cc|c|ccc}
\toprule
                     \multicolumn{1}{c}{\multirow{2}{*}{\textbf{Model}}}    & \multicolumn{1}{c}{\textbf{HoC}}   & \multicolumn{1}{c}{\textbf{PubMedQA}}  & \multicolumn{3}{c}{\textbf{NCBI-Disease}} \\
\cmidrule(lr){2-2} \cmidrule(lr){3-3} \cmidrule(lr){4-6}
             &    \multicolumn{1}{c|}{\textbf{Accuracy}}  & \multicolumn{1}{c|}{\textbf{Accuracy}} & \textbf{Precision} &  \textbf{Recall} & \textbf{F1} \\
\midrule
GPT-4o-Mini                   & \textbf{63.04}     & 55.6          & 20.71         & 21.88      & 21.28 \\
Gemini-1.5-flash               & 55.86     & 54.0          & \textbf{52.94}         & \textbf{49.69}      & \textbf{51.26}\\

Claude-3-Haiku        & 52.48     & \textbf{61.6 }         & 18.54         & 27.29      & 22.08 \\
\midrule
Phi-3.5-Mini-3.8B-Instruct          & 49.45     & \textbf{58.4}          & 6.81          & \textbf{28.23}      & 10.98 \\
Mistral-7B-v0.3-Instruct       & 49.47     & 57.2          & 4.41          & 21.98      & 7.35  \\

Qwen-2.5-7B-Instruct            & \textbf{62.41}     & 23.2          & \textbf{19.29}         & 25.00      & \textbf{21.78} \\
LLaMA-3.1-8B-Instruct     & 14.83     & 55.0          & 8.13          & 13.75      & 10.22 \\
\midrule
DeepSeek-R1-Distill-Qwen-7B    & 49.02     & 54.0          & \textbf{19.71}         & \textbf{27.08}      & \textbf{22.82} \\
DeepSeek-R1-Distill-LLaMA-8B   & \textbf{52.68}     & \textbf{59.6 }         & 10.02         & 23.54      & 14.06 \\

\bottomrule
\end{tabular}

\caption{Results on HoC, PubMedQA, and NCBI-Disease datasets.}
\label{tab:res_text}
\end{table}

\begin{table}[t]
\centering
\tiny
\begin{tabular}{c*{4}{c}|*{4}{c}}
\toprule
 \multicolumn{1}{c}{\multirow{2}{*}{\textbf{Model}}} & \multicolumn{4}{c}{\textbf{MediQA-QS}} & \multicolumn{4}{c}{\textbf{MediQA-ANS}} \\
\cmidrule(lr){2-5} \cmidrule(lr){6-9}
                       & \textbf{R-1} & \textbf{R-2} & \textbf{R-L} & \textbf{B-S} & \textbf{R-1} & \textbf{ R-2} & \textbf{R-L} & \textbf{B-S} \\
\midrule

GPT-4o-Mini                   & 28.79 & 10.95 & 22.36 & 89.15 & 30.14 & 9.26  & 19.15 & 87.09 \\

Gemini-1.5-Flash               & \textbf{33.25} & \textbf{12.50} & \textbf{27.65} & \textbf{89.85} & 28.44 & 8.75  & 19.50 & 86.87 \\
Claude-3-Haiku       & 28.21 & 11.12 & 23.77 & 88.83 & \textbf{31.01} & \textbf{11.45} & \textbf{19.88} & \textbf{86.49} \\
\midrule
Phi-3.5-Mini-3.8B          & \textbf{28.49} & \textbf{10.29} & \textbf{22.89} & \textbf{89.07} & 25.63 & 7.12  & 15.39 & 85.65 \\
Qwen-2.5-7B           & 25.84 & 8.79  & 19.87 & 88.11 & 27.58 & 8.71  & 18.04 & 86.25 \\
Mistral-7B-v0.3       & 24.47 & 8.56  & 20.00 & 88.14 & 29.20 & 10.21 & 18.20 & 86.29 \\
LLaMA-3.1-8B     & 24.15 & 7.76  & 18.58 & 87.37 & \textbf{32.55} & \textbf{13.28} & \textbf{22.11} & \textbf{86.29} \\
\midrule

DeepSeek-R1-Distill-Qwen-7B    & \textbf{23.16} & \textbf{8.94}  & \textbf{18.47} & \textbf{87.64} & 26.29 & 6.69  & 16.26 & 85.94 \\
DeepSeek-R1-Distill-LLaMA-8B   & 14.40 & 4.09  & 11.27 & 85.52 & \textbf{26.38} & \textbf{7.01}  & \textbf{16.49} & \textbf{86.05} \\

\bottomrule
\end{tabular}
\caption{Text Summarization Results. Here, `ROUGE-' is `R-' and `BertScore' is `B-S'.}

\label{tab:res_summ}
\end{table}

\subsection{Performance in Biomedical Text Processing Tasks}

We show the results of different models in HoC, PubMedQA and NCBI-Disease datasets in Table \ref{tab:res_text} and in MediQA-QS and MediQA-ANS datasets in Table \ref{tab:res_summ}. Based on the results, we find that there is not a single LLM that achieves the best result across all datasets. 

For instance, GPT-4o-Mini achieves the best in HoC, whereas Gemini-1.5-Flash and Claude-3-Haiku achieve the best result in NCBI-Disease and PubMedQA datasets, respectively. In summarization, we find that Gemini-1.5-Flash has the best result in MediQA-QS while Claude-3-Haiku outperforming GPT-4o-Mini and Gemini-1.5-Flash in MediQA-ANS. 

In terms of open-source LLMs, we find that they perform on par (and in some cases even better) than closed-source LLMs. For instance, Qwen-2.5-7B-Instruct even outperforms Gemini and Claude in HoC, while Phi-3.5 outperforms GPT-4o and Gemini in PubMedQA. Interestingly, LLaMA-3.1-8B achieves the best result across all models in MediQA-ANS. 

None of the DeepSeek models could achieve the best result in any datasets, although they still achieve decent results. Among the DeepSeek models, we find that the DeepSeek-Distilled model based on Qwen-7B performs better than LLaMA-8B in NCBI-Disease and MediQA-QS,  the opposite happens in HoC, PubMedQA, and MediQA-ANS datasets.   

With the performance of LLMs varying across datasets, LLMs can be chosen for fine-tuning or zero-shot inference based on their task-specific performance in different datasets.

\subsection{Performance in Biomedical Image Processing (Multimodal) Tasks}
We show the results for Molecular Image Captioning and Pathology Image Question Answering (QA) in Table \ref{tab:res_image}. While performance in Molecular Image Captioning is quite similar for most LLMs (except LLaMA-3.2-11B-Vision), many LLMs perform quite poorly in PathVQA, with only Gemini-1.5-Flash and Janus-Pro-7B achieving more than 40\% accuracy. While LLaMA-3.2-11B-Vision performs quite poorly in image captioning, it performs quite better in PathVQA (third best among 8 multimodal models). Among all models,  Janus-Pro-7B and Gemini-1.5-Flash achieve the most consistent results in both datasets, establishing themselves as a good choice for multimodal biomedical tasks. 
\subsection{Model Scaling Experiments}

\begin{figure}
    \centering
  \includegraphics[width=0.85\linewidth]{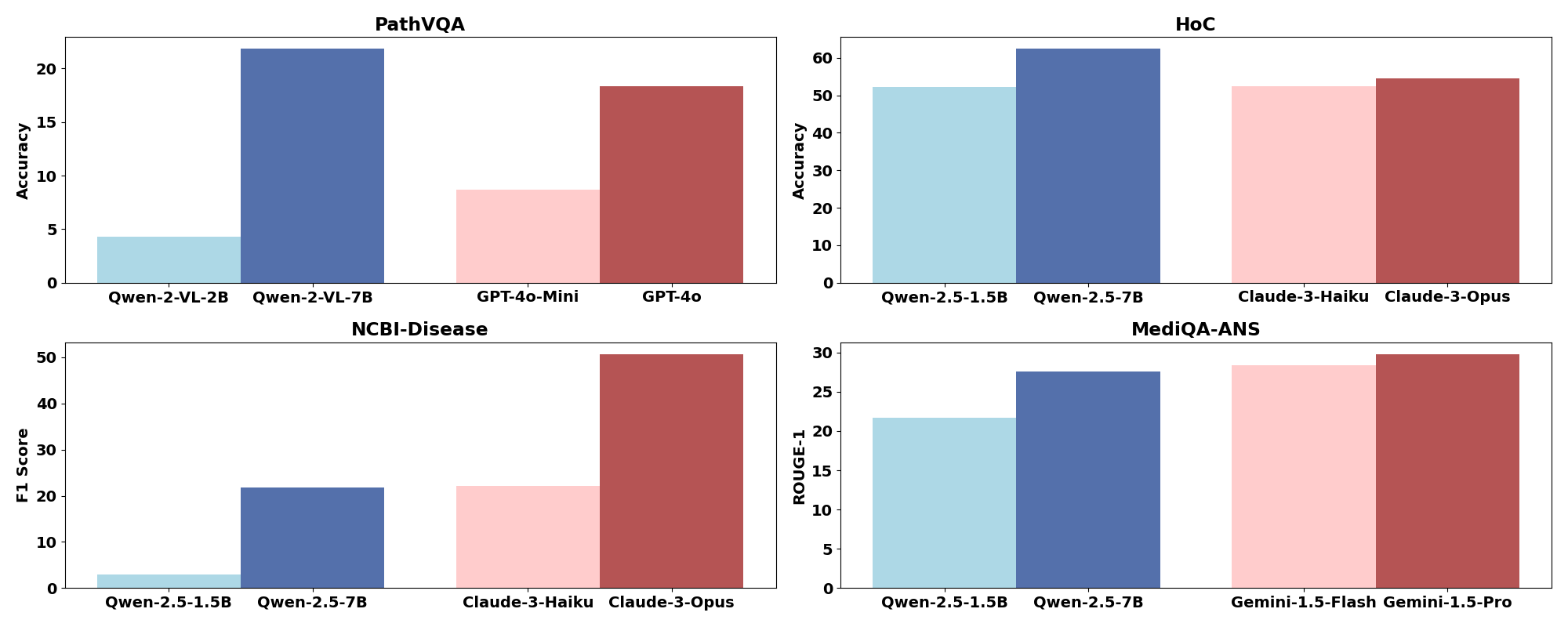}
  
    \caption{Model Scaling Results on \textit{Multimodal} QA (PathVQA), alongside \textit{Text-based} Classification (HoC), NER (NCBI-Disease), and Summarization (MedQA-Ans).}
    \label{fig:scaling}
\end{figure}

In this section, we conduct some model scaling experiments to investigate (i) can scaling up the model size for closed-source LLMs improves the performance (this will give us insights on whether larger closed-source LLMs can be utilized as a better synthetic data generator to train smaller open-source LLMs), and (ii) can scaling down the model size for open-source models retains their performance (this will provide insights on whether more cost-efficient models are reliable in real-world scenarios). For the closed-source LLMs, we select the worst-performing model in the respective dataset (see Figure \ref{fig:scaling}) to investigate their performance with their larger counterpart: Gemini-1.5 (Flash vs Pro), GPT-4 (o-mini vs o), and Claude-3 (Haiku vs Opus). For the open-source LLMs, we compared the Qwen models of various sizes. 
From Figure \ref{fig:scaling}, we find that scaling up the model size is always helpful for the closed-source models, while scaling down leads to a performance drop for open-source models. 

\begin{table}[t]

\centering

\begin{tabular}{lccc|c}

\toprule
 & \multicolumn{3}{c}{\textbf{Molecular Image Captioning}} & \multicolumn{1}{c}{\textbf{PathVQA}} \\
\cmidrule(lr){2-4} \cmidrule(lr){5-5}
Model & \textbf{ROUGE-1} & \textbf{ROUGE-2} & \textbf{ROUGE-L} & \textbf{Accuracy} \\
\midrule
GPT-4o-Mini                 & 19.24 & 2.06 & 12.69 & 8.27  \\

Gemini-1.5-Flash            & 21.61 & 2.64 & 13.11 & \textbf{40.28} \\
Claude-3-Haiku     & \textbf{22.03} & \textbf{2.67} & \textbf{15.01} & 8.49  \\
\midrule
Phi-3.5-Vision-4.2B       & 19.67 & 1.91 & 13.81 & 15.93 \\

Qwen2-VL-7B         & 20.24 & 3.05 & 14.18 & 21.84 \\
LLaVA-Next-7B      & 19.43 & 3.26 & 13.85 & 5.25  \\
Janus-Pro-7B                & \textbf{21.23} & \textbf{3.51} & \textbf{14.29} & \textbf{41.19}  \\
LlaMA-3.2-11B-Vision & 12.69 & 1.88 & 9.52  & 32.25 \\

\bottomrule
\end{tabular}
\caption{{Results for Molecular Image Captioning and Pathology Image QA.}}
\label{tab:res_image}
\end{table}

\section{Conclusion and Future Work}
This study evaluates cost-efficient LLMs across diverse biomedical tasks, covering text and image modalities. With no single model consistently outperforming others, we observe the task-specific nature of existing LLMs in biomedicine. Notably, some open-source models match or surpass closed-source ones while offering efficiency and greater privacy. Our findings guide future research in selecting the right models for further training 
on complex tasks \cite{laskar2025improving}. 
Expanding evaluations to broader biomedical datasets will also enhance our understanding of cost-efficient LLMs in practical healthcare applications \cite{huang2009bayesian}.

\section*{Acknowledgements}
This research was supported by the Natural Sciences and Engineering Research Council (NSERC) of Canada, the York Research Chairs (YRC) program, and Compute Canada. 
 \printbibliography[heading=subbibintoc]

\end{document}